\documentclass[acmtog]{acmart} 
\acmSubmissionID{2122}
\acmJournal{TOG}

\usepackage[ruled,vlined]{algorithm2e} 

\SetAlFnt{\small}
\SetAlCapFnt{\small}
\SetAlCapNameFnt{\small}
\SetAlCapHSkip{0pt}

\usepackage{enumitem}
\usepackage{booktabs}
\usepackage{multirow}
\usepackage{acmart-taps}
\usepackage{graphicx}
\usepackage{caption}
\usepackage{pifont}
\newcommand{\cmark}{\ding{51}}
\newcommand{\xmark}{\ding{55}}
\definecolor{lightred}{RGB}{255,210,210}
\definecolor{lightorange}{RGB}{255,230,180}
\definecolor{lightyellow}{RGB}{255,248,180}

\copyrightyear{2026}
\acmYear{2026}
\setcopyright{cc}
\setcctype{by-nc-nd}
\acmConference[SA Conference Papers '26]{SIGGRAPH Asia 2026 Conference Papers}{December 01--04, 2026}{Kuala Lumpur, Malaysia}
\acmBooktitle{SIGGRAPH Asia 2026 Conference Papers (SA Conference Papers '26), December 01--04, 2026, Kuala Lumpur, Malaysia}
\acmDOI{10.1145/3829340.3842316}
\acmISBN{979-8-4007-2842-6/2026/12}
\begin{document}

\title{OceanXL: Large-scale Underwater 3D Gaussian Splatting via Block Partitioning and Adaptive Pruning}

\author{Haoran Wang}
\affiliation{%
  \institution{University of Bristol}
  \country{UK}
}
\email{yp22378@bristol.ac.uk}

\author{Shaoyu Cai}
\affiliation{%
  \institution{National University of Singapore}
  \country{Singapore}
}
\email{shaoyucai@nus.edu.sg}

\author{Adrian Azzarelli}
\affiliation{%
  \institution{University of Bristol}
  \country{UK}
}
\email{a.azzarelli@bristol.ac.uk}

\author{Zhuodong Jiang}
\affiliation{%
  \institution{University of Bristol}
  \country{UK}
}
\email{zhuodong.jiang@bristol.ac.uk}

\author{Guoxi Huang}
\affiliation{%
  \institution{University of Bristol}
  \country{UK}
}
\email{guoxi.huang@bristol.ac.uk}

\author{Eng Tat Khoo}
\affiliation{%
  \institution{National University of Singapore}
  \country{Singapore}
}
\email{etkhoo@nus.edu.sg}

\author{Brett Seymour}
\affiliation{%
  \institution{National Park Service's Submerged Resources Center}
  \country{USA}
}
\email{Brett_Seymour@nps.gov}

\author{Fan Zhang}
\affiliation{%
  \institution{University of Bristol}
  \country{UK}
}
\email{Fan.Zhang@bristol.ac.uk}

\author{David Bull}
\affiliation{%
  \institution{University of Bristol}
  \country{UK}
}
\email{Dave.Bull@bristol.ac.uk}

\author{Nantheera Anantrasirichai}
\affiliation{%
  \institution{University of Bristol}
  \country{UK}
}
\email{N.Anantrasirichai@bristol.ac.uk}


\begin{abstract}
Underwater 3D reconstruction is critical for marine exploration, ecological monitoring, and subsea infrastructure inspection, yet remains challenging at large scale due to light attenuation, scattering, and limited capture coverage. While 3D Gaussian Splatting (3DGS) enables high-quality real-time rendering, its application to large underwater scenes is constrained by high memory consumption and inefficient optimization over extensive areas. We propose OceanXL, a fast and scalable 3DGS-based framework for large-scale underwater reconstruction. OceanXL adopts a divide-and-conquer strategy, partitioning scenes into spatially coherent blocks to enable efficient optimization while preserving global geometric consistency. We further introduce an adaptive pruning scheme tailored to underwater conditions that removes redundant primitives, producing compact representations without sacrificing visual fidelity. Together, these components improve training efficiency and rendering performance for large scenes. We also introduce a large-scale underwater dataset covering diverse marine environments. Experiments on five large-scale scenes demonstrate favorable scalability, compactness, and efficiency--quality trade-offs over large-scene baselines. Controlled comparisons on the small-scale SeaThru-NeRF dataset further show competitive reconstruction quality with substantially smaller model sizes than underwater-specific methods. Code and datasets are publicly available at \url{https://wanghaoran16.github.io/OceanXL-Large-scale-Underwater-3D-Gaussian-Splatting/}.

\end{abstract}



\begin{teaserfigure}
  %  \fbox{\rule{0pt}{1.7in} \rule{0.98\linewidth}{0pt}}
  \includegraphics[width=0.95\textwidth]{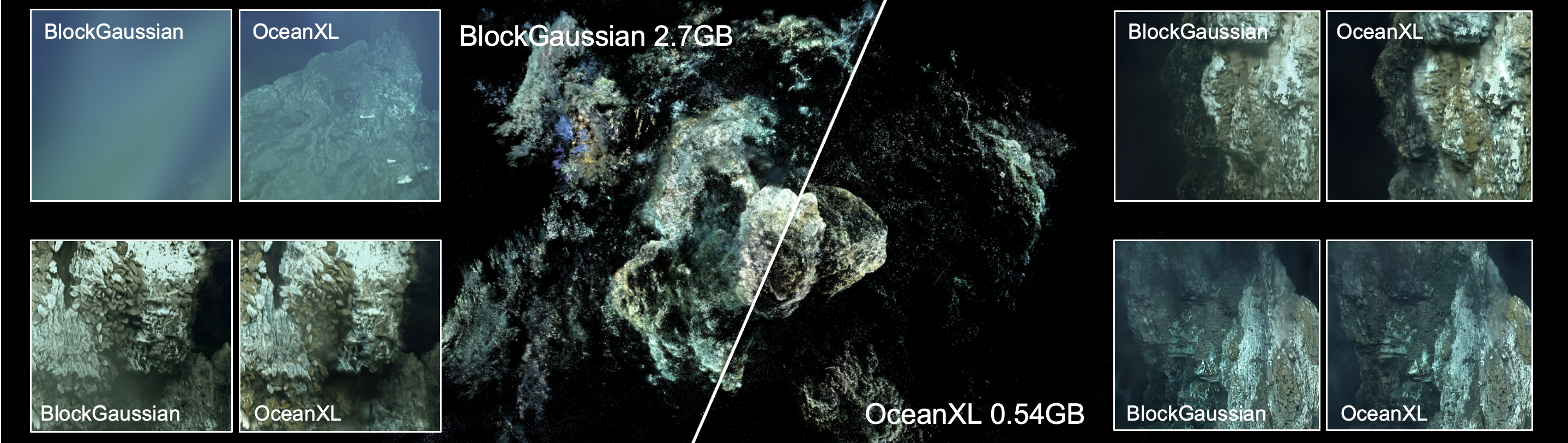}
  \caption{\textbf{OceanXL}: A scalable framework for large-scale underwater 3D Gaussian Splatting that achieves compact scene representations while preserving high-fidelity rendering quality. Compared with the strongest baseline, BlockGaussian~\cite{wu2025blockgaussian}, OceanXL produces more efficient reconstructions with improved visual quality across diverse underwater environments. Example shown: \textbf{Eiffel Tower 2016}.}
  \Description{Large-scale underwater scene reconstruction}
  \label{fig:teaser}
\end{teaserfigure}

\maketitle

\section{Introduction}
High-fidelity three-dimensional (3D) reconstruction of underwater environments underpins a wide range of applications, including marine habitat monitoring \cite{Marre:monitor:2019,Zhong2025HighQuality}, underwater archaeology and infrastructure inspection \cite{Tanduo:underwater:2025}, and immersive interactive experiences \cite{ouyang2025oceanvive}. More recently, digital twins of shipwrecks, coral reefs, and deep-sea environments have become increasingly important for virtual reality (VR), augmented reality (AR), film production, and interactive storytelling \cite{thomas2018oceans}. The ability to author and explore such scenes in real time also enables new forms of Human–Computer Interaction (HCI), including virtual dive simulations \cite{jain2016immersive}, educational exhibits \cite{fauville2025ocean}, and collaborative ocean exploration \cite{maggipinto2025diving}. 

Reconstruction-driven asset capture approaches, such as photogrammetry and multi-view stereo (MVS), have become standard techniques for generating detailed 3D content from image sequences. However, underwater environments remain particularly challenging for image-based 3D reconstruction due to light absorption, scattering, wavelength-dependent color attenuation, and non-uniform illumination. These effects degrade image quality, distort geometry and appearance consistency, and introduce artifacts that reduce the visual fidelity required for cinematic rendering and immersive experiences. The problem becomes more severe in large-scale underwater scenes, where limited visibility and uneven lighting often produce incomplete geometry, noisy reconstructions, and inconsistent appearance across viewpoints.

Recent advances in neural rendering have significantly improved 3D reconstruction. Among them, Neural Radiance Fields (NeRF) \cite{mildenhall2021nerf} demonstrate impressive scene representation quality but remain limited by slow training and rendering speeds, particularly for large-scale environments. To address these limitations, 3D Gaussian Splatting (3DGS)~\cite{kerbl20233d} has emerged as an efficient alternative for real-time neural scene representation. By modeling scenes with anisotropic Gaussian primitives optimized directly in 3D space, 3DGS achieves high-quality rendering with substantially improved training and rendering efficiency, making it well suited for novel-view synthesis, asset capture, and immersive ocean exploration.

Building upon this success, several recent studies have extended Gaussian-based rendering methods to underwater environments \cite[e.g.,][]{wang2025uw,li2024watersplatting,jiang2025rusplatting,zhang2025recgs,Yi:AtlantisGS:2025}. These approaches introduce underwater-aware rendering models or incorporate physical priors to better account for underwater light propagation, improving reconstruction quality under scattering, attenuation, and visibility degradation. In parallel, recent work on compact Gaussian representations \cite[e.g.,][]{fang2024mini,niemeyer2025radsplat,wang2026prune} has explored adaptive pruning and primitive refinement strategies to reduce redundancy in Gaussian primitives and improve model compactness. 

Despite these advances, existing methods remain limited when applied to large-scale underwater environments. Most underwater Gaussian splatting approaches are designed for relatively small scenes~\cite{levy2023seathru,wang2025uw} and do not address the scalability requirements of large immersive environments for VR/AR applications. At the same time, compact Gaussian strategies have rarely been investigated under underwater degradation, where scattering, turbidity, and color distortion destabilize geometry estimation and radiance modeling. In large underwater scenes, representing spatially varying visibility and attenuation often requires a substantial number of Gaussian primitives, significantly increasing memory footprint and slowing both training and rendering throughput, limiting the real-time and interactive requirements of immersive applications. Moreover, underwater imaging inconsistencies can hinder effective pruning, as noisy or biased observations may be incorrectly preserved as valid structures. As a result, efficient and scalable reconstruction for large underwater environments remains largely underexplored.

Motivated by these challenges, we propose OceanXL, a fast and scalable framework for large-scale underwater reconstruction based on 3D Gaussian Splatting. OceanXL adopts a divide-and-conquer strategy that partitions underwater scenes into spatial blocks for efficient optimization. Each block is reconstructed independently and then merged into a coherent global representation. To further improve efficiency, we introduce an underwater-aware adaptive pruning mechanism that removes redundant Gaussian primitives, reducing model complexity while preserving reconstruction fidelity. As illustrated in Figure~\ref{fig:teaser}, OceanXL achieves compact, high-fidelity reconstruction of large underwater environments while improving efficiency over existing large-scene Gaussian splatting methods.

Existing underwater reconstruction datasets are largely limited to small-scale scenes, controlled capture settings, or localized object-centric reconstructions, restricting the study of scalable neural rendering for immersive underwater exploration. Large-scale datasets suitable for evaluating modern neural scene representations remain scarce, particularly for environments with extensive spatial coverage, severe visibility degradation, complex coral structures, and long unconstrained capture trajectories. To address this gap, we introduce large-scale underwater datasets designed to support 3D neural reconstruction for immersive ocean experiences. The datasets cover diverse underwater environments with extensive spatial coverage, enabling systematic evaluation of scalable reconstruction methods. We hope this work will inspire future neural rendering systems for visually complex underwater environments and broader applications in immersive graphics.

The main contributions can be summarized as follows:
\begin{enumerate}[leftmargin=*]
    \item We present OceanXL, a scalable underwater 3D Gaussian Splatting framework that combines adaptive scene decomposition with compact Gaussian optimization for efficient large-scale underwater reconstruction.
    \item We introduce an underwater-aware lightweight density control strategy that compensates for attenuation-induced densification failures while progressively pruning redundant primitives, improving scalability without sacrificing reconstruction fidelity.
    \item We propose a selective Difference-of-Gaussians (DoG) activation mechanism that enhances local representational capacity only in reconstruction-sensitive regions, enabling compact yet high-fidelity underwater scene representations.
    \item We introduce large-scale underwater datasets spanning shipwrecks, coral reefs, and deep-sea environments, together with extensive evaluations on large-scale scenes demonstrating favorable trade-offs between scalability, efficiency, and reconstruction quality, complemented by SeaThru-NeRF comparisons demonstrating compactness against underwater-specific methods.
\end{enumerate}
%\end{itemize}

%Extensive experiments demonstrate that the proposed method significantly improves reconstruction efficiency while maintaining high visual quality. Compared with baseline Gaussian splatting approaches, our framework achieves faster training, more compact scene representations, and improved scalability for large underwater environments.

% --------------------------------------------------
\section{Related Work}
\label{sec:relatedwork}

\subsection{Underwater 3DGS Reconstruction}
Extending 3DGS to underwater environments has gained attention but remains challenging due to scattering, attenuation, and dynamic disturbances~ \citep{huang:Visual:2025}, which violate the standard assumptions of consistent radiance and visibility. Existing efforts primarily adapt 3DGS by incorporating underwater image formation models and physical priors. For instance, UW-GS~\citep{wang2025uw} introduces physics-aware density control and auxiliary supervision to mitigate scattering and motion artifacts, while WaterSplatting~\citep{li2024watersplatting} explicitly models medium transmittance to improve visual realism. Aquatic-GS~\citep{liu2024aquatic} further combines implicit water representations with explicit Gaussians for more physically consistent reconstruction. More recent approaches explore enhanced robustness under degraded observations: SeaSplat~\citep{yang2025seasplat} disentangles scene radiance from medium effects, RecGS~\citep{zhang2025recgs} employs recurrent refinement to stabilise reconstruction, and RUSplatting~\citep{jiang2025rusplatting} incorporates uncertainty modeling to handle noisy inputs. R-Splatting~\citep{huang2025fromrestoration} integrates image enhancement priors to compensate for illumination and color degradation. AtlantisGS~\citep{Yi:AtlantisGS:2025} and SWAGSplatting \cite{jiang_semantic2026} separate foreground content from the background medium and reallocate primitives to salient regions, improving reconstruction quality under sparse observations.

\subsection{Pruning Gaussian Splats}
Despite the rapid progress of 3DGS, improving computational efficiency remains a major challenge. Standard density control strategies often introduce redundant primitives, increasing memory usage and optimization cost~\cite{bagdasarian20253dgs}. To address this, many recent works focus on pruning strategies that retain only the most informative Gaussians. Early approaches estimate importance using opacity~\cite{navaneet2024compgs, zhang2025gaussianspa}, while later methods adopt more targeted metrics based on accumulated ray contributions~\cite{niemeyer2025radsplat}, blending weights~\cite{fang2024mini}, spatial sensitivity~\cite{hanson2025pup}, or per-Gaussian gradients~\cite{hanson2025speedy}. Other approaches, such as MaskGaussian~\cite{liu2025maskgaussian} and LP-3DGS~\cite{zhang2024lp}, further employ differentiable masking strategies to learn adaptive importance weights.
While these methods improve compactness and efficiency, they are primarily designed for in-air scenes with relatively clean observations. Their effectiveness under severe underwater degradation and spatially varying visibility remains largely unexplored.

\subsection{3DGS-based Large Scene Reconstruction}
Recent advances in scalable 3DGS address efficiency and large-scale reconstruction through hierarchical designs and system optimization. Divide-and-conquer approaches partition scenes spatially: Hierarchical-GS \cite{kerbl2024hierarchical} decomposes scenes into hierarchical chunks, enabling joint optimization and level-of-detail (LoD) control, while CityGaussian~\cite{liu2024citygaussian} employs block-wise training with compression-based LoD generation. BlockGaussian~\cite{wu2025blockgaussian} introduces visibility-aware optimization and pseudo-view constraints to reduce block-merging artifacts. Structured LoD representations organize primitives hierarchically: Octree-GS~\cite{ren2024octree} uses octree-based multi-scale organization with grow-and-prune densification, whereas MixGS~\cite{liu2025holistic} integrates poses and attributes into view-aware latent representations decoded into fine-scale primitives. System-level optimizations target computational bottlenecks: FlashGS~\cite{Feng_2025_CVPR} reduces redundant computation through precise intersection tests and adaptive scheduling, while GS-Scale~\cite{Lee:GSScale:2026} addresses memory constraints via host-GPU transfers, selective offloading, and deferred updates to minimize overhead.
% --------------------------------------------------

%\vspace{-4mm}
\section{OceanXL: Fast Large-scale Underwater Representation}
To enable large-scale underwater scene reconstruction, we propose OceanXL, an efficient divide-and-conquer pipeline that decomposes extensive scenes into manageable subproblems, reconstructs them independently, and integrates them into a unified representation.

Starting from a sparse point cloud and camera poses estimated via SfM, OceanXL first applies the proposed Balanced Scene Partitioning (BSP) scheme (\autoref{ssec:BSP}) to generate spatially balanced partitions by adaptively adjusting sub-scene sizes and distributing optimization complexity across chunks. Each partition is then reconstructed using the proposed Underwater-Pruning Accelerated Gaussian Splatting framework (\autoref{ssec:pruning}), which combines lightweight underwater-aware density control with a compact 3D Difference-of-Gaussians (DoG) representation. This design reduces model complexity and optimization cost while preserving reconstruction fidelity. Finally, all reconstructed partitions are merged into a coherent global representation.

\begin{figure*}
    \includegraphics[width=\linewidth,trim=5mm 2mm 6mm 2mm,clip]{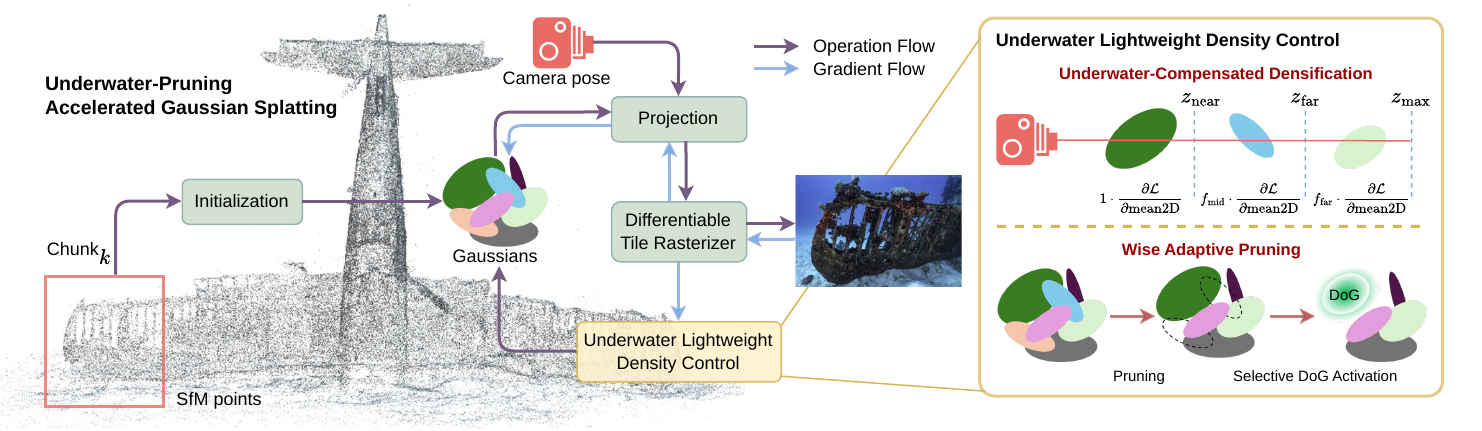} %\vspace{-4mm}
    \caption{Overview of the proposed \textbf{Underwater-Pruning Accelerated Gaussian Splatting} pipeline. Starting from SfM points of chunks generated by \textbf{Balanced Scene Partitioning}, Gaussian primitives are initialized and projected onto the 2D image plane for fast rasterization. The proposed \textbf{Underwater Lightweight Density Control} module, highlighted in yellow, regulates primitives through two components: \textbf{Underwater-Compensated Densification}, which uses a few learnable factors to compensate for densification failures caused by underwater light attenuation, and \textbf{Wise Adaptive Pruning}, which removes redundant primitives while progressively introducing 3D-DoG components for high-frequency detail reconstruction.
}
\vspace{-4mm}
\Description{Overview of the OceanXL pipeline, including balanced scene partitioning, underwater-compensated densification, adaptive pruning, and 3D Difference-of-Gaussians components.}
    \label{fig:pipeline}
\end{figure*}
% -------------------------------------------
\begin{algorithm}[!t]
\caption{Balanced Scene Partitioning}
\label{alg:balanced_partition}

\KwIn{Sparse SfM points $\mathcal{P}$, cameras $\mathcal{C}$, point threshold $N_p$, maximum depth $D$, visibility threshold $\tau$}
\KwOut{Local reconstruction chunks $\{\mathcal{S}_k\}$}

Compute global bounding box $\mathcal{B}$ of $\mathcal{P}$\;
Initialize queue $\mathcal{Q} \leftarrow \{(\mathcal{B}, 0)\}$ and chunk set $\mathcal{R} \leftarrow \emptyset$\;

\While{$\mathcal{Q}$ is not empty}{
    Pop $(\mathcal{B}_k, d)$ from $\mathcal{Q}$\;
    Count points $\mathcal{P}_k \leftarrow \{p \in \mathcal{P} \mid p \in \mathcal{B}_k\}$\;

    \If{$|\mathcal{P}_k| > N_p$ \textbf{and} $d < D$}{
        Find longest spatial axis of $\mathcal{B}_k$\;
        Sample candidate split ratios along this axis\;
        Compute point counts $N_1$ and $N_2$ for each candidate split\;
        Discard degenerate splits with $N_1=0$ or $N_2=0$\;
        Select the split that minimizes
        $\frac{|N_1-N_2|}{\max(N_1,N_2)}$\;

        Split $\mathcal{B}_k$ into two child chunks $\mathcal{B}_{k}^{1}$ and $\mathcal{B}_{k}^{2}$\;
        Push $(\mathcal{B}_{k}^{1}, d+1)$ and $(\mathcal{B}_{k}^{2}, d+1)$ into $\mathcal{Q}$\;
    }
    \Else{
        Add $\mathcal{B}_k$ to $\mathcal{R}$\;
    }
}

\ForEach{chunk $\mathcal{B}_k \in \mathcal{R}$}{
    Expand $\mathcal{B}_k$ to obtain an enlarged chunk region $\hat{\mathcal{B}}_k$\;
    Collect sparse points $\mathcal{P}_k \leftarrow \{p \in \mathcal{P} \mid p \in \hat{\mathcal{B}}_k\}$\;

    \If{$|\mathcal{P}_k|$ is too small}{
        Discard this chunk\;
        \textbf{continue}\;
    }

    Select cameras with sufficient SfM point overlap with $\hat{\mathcal{B}}_k$\;
    Classify selected cameras into base and border views based on overlap ratios\;
    Add extra sparse points observed by the selected cameras\;

    \If{the number of selected cameras is too small}{
        Discard this chunk\;
        \textbf{continue}\;
    }

    Export the valid chunk as a local reconstruction unit $\mathcal{S}_k$\;
}
\Return{$\{\mathcal{S}_k\}$}
\end{algorithm} 
\subsection{Balanced Scene Partitioning}
\label{ssec:BSP}

To address uneven geometry and view distributions in large-scale underwater scenes, we propose a Balanced Scene Partitioning (BSP) strategy that adaptively decomposes the scene, rather than relying on fixed-size grids as in \cite{kerbl2024hierarchical, liu2024citygaussian}. Fixed-grid partitioning often yields imbalanced partitions: dense regions contain an excessive number of points and camera observations, while sparse regions contribute little useful information. This imbalance reduces computational efficiency and leads to inconsistent optimization costs across sub-scenes.

OceanXL mitigates this issue through content-aware partitioning. Given the sparse SfM point cloud and estimated camera poses, we first compute the scene bounding box. To automatically determine the partitioning dimensionality, we compute the point-cloud extents along the three principal axes, denoted by $l_1 \geq l_2 \geq l_3$. We use 2D BSP when $l_3/l_2 < \tau_{\mathrm{geo}}$, indicating an approximately planar scene; otherwise, 3D BSP is applied. The same threshold $\tau_{\mathrm{geo}}=0.3$ is used for all scenes. We then recursively split a partition whenever its point count exceeds a predefined threshold, with each split performed along the longest spatial axis~\cite{Kay:Ray:1986}. Unlike~\cite{wu2025blockgaussian}, which partitions at the geometric center, we adaptively adjust the splitting boundary to produce two child partitions with approximately balanced point counts, yielding a more uniform computational load. Specifically, we perform a discrete search over candidate split ratios $\mathcal{R}$ along the longest axis. For each $r \in \mathcal{R}$, the point cloud is divided into two child partitions with sparse point counts $N_1(r)$ and $N_2(r)$. The optimal split ratio $r^*$ is selected by minimizing the relative imbalance:

{
\begin{equation}
   r^* = \arg\min_{r \in \mathcal{R}} \frac{|N_1(r)-N_2(r)|}{\max(N_1(r),N_2(r))},
\end{equation}
}
where $\mathcal{R}$ denotes the set of valid sampled ratios within $[0.3,0.7]$. Unlike traditional graphics applications that primarily optimize traversal efficiency, our formulation is designed for sparse point clouds in the context of learning-based reconstruction, where balanced geometry and view distributions are important for stable optimization and reconstruction fidelity. To improve robustness, we further introduce guard rails to avoid degenerate partitions with insufficient geometry or camera coverage. The overall procedure is summarized in~\autoref{alg:balanced_partition}, with representative results shown in~\autoref{fig:chunking}. 

After partitioning, cameras are assigned to each partition based on visibility. Following~\cite{wu2025blockgaussian}, a camera is selected if either a sufficient proportion of its observed SfM points falls inside the partition, or the partition is sufficiently covered by the points observed by this camera. To improve cross-boundary consistency, we include sparse points outside the chunk if they are observed by the selected cameras. Finally, we discard chunks with too few sparse points or too few associated cameras, and export each valid partition as an independent local reconstruction unit for processing.

% --------------------------------------------
\subsection{Underwater-Pruning Accelerated Gaussian Splatting}
\label{ssec:pruning}
To improve large-scale underwater reconstruction, we propose \textbf{Underwater-Accelerated Gaussian Splatting}, which integrates a novel \textbf{Underwater Lightweight Density Control} scheme. The proposed scheme enhances densification under underwater degradation and adaptively prunes
redundant Gaussians to generate compact models for individual partitions. This reduces memory consumption and computational overhead while preserving reconstruction fidelity. The details of the overall pipeline are shown in~\autoref{fig:pipeline}.

Compared with scenes captured in clear media, underwater environments exhibit stronger degradation due to wavelength-dependent absorption and scattering. These effects attenuate image contrast and suppress structural cues, causing standard 3DGS optimization to underestimate scene complexity, particularly in distant regions~\cite{wang2025uw}. Recent underwater Gaussian Splatting methods incorporate physics-based underwater imaging models, commonly based on the revised underwater image formation model~\cite{akkaynak2018revised}, together with auxiliary neural networks to estimate water-medium parameters and compensate for attenuation during densification~\cite{li2024watersplatting,jiang2025rusplatting,liu2024aquatic,jiang_semantic2026}. Additional details of the underwater imaging formulation are provided in the supplementary material (Sec. S1).

As observed in UW-GS~\cite{wang2025uw}, attenuation significantly suppresses the 2D positional gradients used for Gaussian densification, leading to insufficient primitive allocation and under-reconstruction in degraded regions. While existing approaches address this issue using MLP-based parameter estimation, the additional networks introduce non-negligible computational overhead and slow down optimization.
To overcome this limitation, we replace the MLP-based gradient compensation with a novel lightweight, learnable depth-aware densification scheme, termed \textbf{Underwater-compensated Densification}. Specifically, Gaussians are grouped into near, mid, and far ranges based on their camera-space depth. For the mid and far groups, each of them is assigned a learnable densification weight, while the near region is assumed to have attenuation that can be ignored. Given the camera-space depth $z_i$ of Gaussian $i$, we assign it to a depth group and define its densification weight as
\begin{equation}
w_i =
\begin{cases}
1, & z_i < z_{\mathrm{near}}, \\
f_{\mathrm{mid}}, & z_{\mathrm{near}} \leq z_i < z_{\mathrm{far}}, \\
f_{\mathrm{far}}, & z_i \geq z_{\mathrm{far}},
\end{cases}
\label{eq:depth_weight}
\end{equation}
where $f_{\mathrm{mid}}$ and $f_{\mathrm{far}}$ are learnable parameters for compensating underwater attenuation in the mid- and far-depth regions, respectively. Since densification is non-differentiable, $f_{\mathrm{mid}}$ and $f_{\mathrm{far}}$ are optimized using gradient-free SPSA, enforcing
$f_{\mathrm{far}}\geq f_{\mathrm{mid}}$. ~\autoref{fig:pipeline} (top-right) illustrates the proposed scheme.
This design directly modulates densification strength across depth, enabling adaptive emphasis on distant and heavily degraded regions. Importantly, it avoids the need for an auxiliary network, thereby reducing computational overhead while preserving effective geometric reconstruction in underwater scenes.

%Despite its effectiveness, enhanced densification inevitably increases the number of Gaussians, exacerbating the computational and memory burden in large-scale underwater reconstruction. To address this, we adopt a compact Gaussian Splatting framework inspired by~\cite{wang2026prune}, which aims to reduce storage and rendering cost while maintaining reconstruction fidelity. The approach progressively prunes redundant or low-contribution primitives to obtain a more efficient representation. However, aggressive pruning can degrade reconstruction quality, particularly in regions with fine details or high-frequency structures. To compensate for this loss, Difference-of-Gaussian (DoG) primitives is introduced:

To ensure scalability to large scenes, we integrate the proposed densification strategy within a compact Gaussian Splatting framework and propose a \textbf{Wise Adaptive Pruning} scheme. Rather than allowing densification to inflate model complexity, the framework maintains efficiency by progressively pruning redundant or low-contribution primitives, thereby reducing storage and rendering cost. To retain reconstruction fidelity under this compact model, we incorporate Difference-of-Gaussians (DoG) primitives, inspired by~\cite{wang2026prune}. These enhance local representational capacity, enabling the model to preserve fine structures and geometric detail even with fewer Gaussians. The DoG formulation is defined as:
\begin{equation}
  DoG(x) = G(x) - G_{p}(x),
  \label{eq:important}
\end{equation}
where $G_{p}(x)$ denotes a pseudo-Gaussian that shares the same center coordinates and kernel parameters as the primary Gaussian $G(x)$, while differing only in opacity and scale. This DoG formulation increases the local expressive ability of the remaining primitives and helps recover sharp structures after pruning. Thus, the original method combines adaptive pruning and DoG scheme to achieve a trade-off between model compactness and rendering fidelity.

Unlike the original strategy in~\cite{wang2026prune}, which activates DoG components for all primitives and then degenerates insensitive ones back to standard Gaussians, our {Wise Adaptive Pruning} method introduces a \textbf{selective DoG activation strategy}. We observe that global DoG activation introduces unnecessary computational overhead, since many primitives do not contribute significantly to high-frequency reconstruction. Instead, we identify reconstruction-sensitive primitives based on their gradient responses. Specifically, we use the magnitude of the unweighted 2D positional gradient as an indicator of reconstruction sensitivity. A primitive activates its DoG components only when its gradient magnitude exceeds a predefined threshold $\tau$. Formally, the DoG activation indicator for Gaussian $i$ is defined as
\begin{equation}
a_i =
\mathbb{I}\left(
\left\|
\frac{\partial \mathcal{L}}{\partial \mathrm{mean2D}_i}
\right\|_2
> \tau
\right),
\label{eq:dog_activation}
\end{equation}
The gradient in~\autoref{eq:dog_activation} is not weighted by $w_i$ in~\autoref{eq:depth_weight}, as it is used only for densification and would otherwise introduce depth-biased DoG activation. Here, $a_i=1$ activates the DoG component,, while $a_i=0$ retains the standard Gaussian representation. This selective activation avoids redundant computation while preserving representational capacity in detail-critical regions. The benefits of our 3D-DoG representation are clearly demonstrated in ~\autoref{fig:DoG}.

% --------------------------------------------------

\section{Datasets}

Large-scale underwater datasets remain scarce due to the difficulty and cost of underwater data acquisition. Furthermore, synthesizing realistic underwater imagery is considerably more challenging than in many other domains~\cite{li2017watergan, barbosa2025physically}, requiring the modeling of light absorption, scattering, spatially varying turbidity, non-uniform illumination, and dynamic water disturbances. To support the evaluation of large-scale underwater reconstruction, we introduce two complementary datasets: \textbf{Abyssal} (image-based) and \textbf{OceanXplore} (video-based). Representative samples are shown in ~\autoref{fig:Abyssal-Artifacts}.

\textbf{Abyssal dataset} contains two sequences featuring distinctive man-made objects: a shipwreck and an aircraft wreck. The sequences were captured by the Submerged Resources Center of the National Park Service and are denoted as \emph{ISRO} and \emph{KWAJ}, respectively. \emph{ISRO} contains 2,239 images of the Tokai Maru shipwreck near Guam, captured using a Nikon Z7 through the NPS SeaArray photogrammetry system~\cite{jmse8110849}. \emph{KWAJ} contains 1,294 images of the Kwajalein Atoll F4U-1 Corsair wreck near Mellu Island, captured using a Nikon D5 with a Nauticam housing, a 14--24 mm f/2.8 Nikkor lens, and a 230 mm dome port. The 8K images exhibit some distortion due to low-light conditions and sparse viewpoints. Cleaned photogrammetry models with water removed are provided alongside both sequences. \textbf{OceanXplore dataset} comprises three large-scale coral reef video sequences provided by OceanX, a non-profit ocean exploration organization, denoted as \emph{Komodo}, \emph{Amirantes1}, and \emph{Amirantes2}. \emph{Komodo} was captured using a GoPro near Komodo, Indonesia, at 2704$\times$1520 resolution and 24 fps for 60.77 seconds. It features clear water and abundant dynamic marine life, making it representative of structured yet dynamic reef environments. \emph{Amirantes1} and \emph{Amirantes2} were recorded using a Lumix GH5 near the Amirantes region in the Seychelles at 1920$\times$1080 and 24 fps, lasting 56.31 and 73.82 seconds, respectively. \emph{Amirantes1} exhibits strong illumination variation caused by surface-driven lighting, whereas \emph{Amirantes2} contains dense sea-fan structures with substantial geometric complexity.

The current \textbf{OceanXplore} release is designed primarily to evaluate large-scale reconstruction scalability rather than exhaustively cover underwater environments. The three sequences were selected from a larger archive for their extensive spatial coverage and predominantly static scene structure, although transient fish motion remains, particularly in \emph{Komodo}. Upon acceptance, we will release the sequences together with extracted frames, camera poses, evaluation splits, and preprocessing scripts, and expand the collection in future releases.

In addition to the newly introduced datasets, we further evaluate OceanXL on the large-scale public \emph{Eiffel Tower} dataset~\cite{boittiaux2023eiffel} and \emph{Tabuhan P1}, a subset of the Sweet Corals dataset~\cite{wildflow_sweet_corals_indo_2025}. 

\section{Experiments}
%\subsection{Experimental Setup}

\subsection{Baselines and Metrics} 
We compare our method against several large-scale 3DGS baselines, including Hierarchical-GS~\cite{kerbl2024hierarchical}, CityGaussian~\cite{liu2024citygaussian}, Octree-GS~\cite{ren2024octree}, and BlockGaussian~\cite{wu2025blockgaussian}. We additionally combine UW-GS~\cite{wang2025uw} with the proposed BSP strategy to form an underwater-specific baseline. Evaluation includes PSNR, SSIM, and LPIPS, together with model size and training time to assess scalability for large-scale underwater reconstruction.

%\vspace{1mm}
\subsection{Implementation}
We employ COLMAP~\cite{schonberger2016structure} to initialize the point cloud and estimate camera poses. The partitioning dimensionality is automatically selected using the criterion in~\autoref{ssec:BSP}. For ISRO and Tabuhan P1, we use 2D partitioning by projecting points onto the ground plane, following common large-scale reconstruction pipelines~\cite{liu2024citygaussianv2,gao2025citygs,kerbl2024hierarchical}. For KWAJ and Eiffel Tower 2016, which contain stronger vertical structures, we adopt 3D partitioning by recursively splitting the sparse point cloud along the longest axis of its 3D bounding box. This approach is more appropriate for scenes with vertical structure and produces more balanced partitions in highly elevated environments.

The proposed method is built upon 3DGS~\cite{kerbl20233d}.
After partitioning, each chunk is optimized for 40K iterations. The point cloud is densified in the first 20K iterations, after that, we adopt a modified pruning process with 3D-DoGs introduction for another 10,000 iterations and only retain 50\% of the original primitives. The final model is produced by merging individual chunks. Compared with other methods, our approach can be trained with limited computational resources, using only a single RTX3090 GPU.

\aptLtoX{\begin{table*}[t]
    \setlength\tabcolsep{2pt}
    \captionsetup{singlelinecheck=false}
    \caption{Quantitative evaluation of the proposed approach compared with previous baselines on five large-scale underwater scenes: ISRO, KWAJ, Eiffel Tower 2016, Tabuhan P1, and Komodo. $\uparrow$ indicates higher is better, while $\downarrow$ indicates lower is better. Model size is reported in GB. The \colorbox{lightred}{best}, \colorbox{lightorange}{second best}, and \colorbox{lightyellow}{third best} results are highlighted. UW-GS$^{\ast}$ denotes a modified version combined with BSP. OOM indicates out-of-memory failure.}   
    \begin{tabular}{r | ccccr | ccccr | ccccr | ccccr | ccccr}
        \hline
        \multirow{2}{*}{Method}
        & \multicolumn{5}{c}{ISRO}
        & \multicolumn{5}{c}{KWAJ}
        & \multicolumn{5}{c}{Eiffel Tower 2016}
        & \multicolumn{5}{c}{Tabuhan P1}
        & \multicolumn{5}{c}{Komodo} \\
%        \cmidrule(l{2pt}r{2pt}){2-6} 
%\cmidrule(l{2pt}r{2pt}){7-11} 
%\cmidrule(l{2pt}r{2pt}){12-16} 
%\cmid(l{2pt}r{2pt}){17-21} 
%\cmidrule(l{2pt}r{2pt}){22-26}

\cmidrule(lr){2-6} \cmidrule(lr){7-11} \cmidrule(lr){12-16} \cmidrule(lr){17-21} \cmidrule(lr){22-26}

        & Size$\downarrow$ & PSNR$\uparrow$ & SSIM$\uparrow$ & LPIPS$\downarrow$ & Time$\downarrow$
        & Size$\downarrow$ & PSNR$\uparrow$ & SSIM$\uparrow$ & LPIPS$\downarrow$ & Time$\downarrow$
        & Size$\downarrow$ & PSNR$\uparrow$ & SSIM$\uparrow$ & LPIPS$\downarrow$ & Time$\downarrow$
        & Size$\downarrow$ & PSNR$\uparrow$ & SSIM$\uparrow$ & LPIPS$\downarrow$ & Time$\downarrow$
        & Size$\downarrow$ & PSNR$\uparrow$ & SSIM$\uparrow$ & LPIPS$\downarrow$ & Time$\downarrow$ \\
        \hline

        UW-GS$^{\ast}$ (+BSP)
        & \colorbox{lightorange}{0.923} & 20.53 & 0.669 & 0.442 & 245m45s
        & 1.070 & 21.32 & 0.677 & 0.448 & 198m22s
        & 1.320 & 19.98 & 0.592 & 0.514 & 342m17s
        & OOM & OOM & OOM & OOM & OOM
        & 0.891 & 22.56 & 0.634 & 0.199 & 87m43s \\

        Hierarchical-GS
        & 13.310 & 20.28 & 0.648 & 0.452 & 320m8s
        & 6.280 & 21.42 & 0.679 & \colorbox{lightyellow}{0.380} & 115m18s
        & 9.110 & 17.30 & 0.519 & 0.462 & \colorbox{lightyellow}{180m56s}
        & 39.530 & 16.28 & \colorbox{lightyellow}{0.404} & \colorbox{lightyellow}{0.395} & \colorbox{lightyellow}{561m50s}
        & 1.590 & 23.98 & 0.796 & 0.184 & 55m51s \\

        CityGaussian
        & 2.050 & 21.42 & 0.695 & 0.490 & \colorbox{lightyellow}{150m10s}
        & 1.090 & 21.90 & 0.699 & 0.405 & 109m27s
        & \colorbox{lightorange}{0.450} & 21.91 & \colorbox{lightyellow}{0.700} & \colorbox{lightyellow}{0.425} & 426m0s
        & \colorbox{lightred}{3.800} & \colorbox{lightyellow}{18.31} & \colorbox{lightyellow}{0.568} & 0.417 & \colorbox{lightorange}{430m12s}
        & 2.440 & \colorbox{lightyellow}{28.09} & \colorbox{lightyellow}{0.905} & \colorbox{lightyellow}{0.163} & 432m49s \\

        Octree-GS
        & \colorbox{lightyellow}{1.420} & \colorbox{lightyellow}{24.14} & \colorbox{lightred}{0.753} & \colorbox{lightred}{0.377} & \colorbox{lightorange}{52m17s}
        & \colorbox{lightorange}{0.373} & \colorbox{lightyellow}{24.20} & \colorbox{lightyellow}{0.724} & \colorbox{lightyellow}{0.358} & \colorbox{lightorange}{37m41s}
        & \colorbox{lightred}{0.184} & \colorbox{lightyellow}{23.20} & 0.671 & 0.444 & \colorbox{lightred}{26m22s}
        & OOM & OOM & OOM & OOM & OOM
        & \colorbox{lightorange}{0.181} & \colorbox{lightorange}{28.31} & \colorbox{lightorange}{0.921} & \colorbox{lightorange}{0.149} & \colorbox{lightyellow}{52m22s} \\

        BlockGaussian
        & 3.300 & \colorbox{lightorange}{24.43} & \colorbox{lightyellow}{0.729} & \colorbox{lightorange}{0.404} & 269m4s
        & \colorbox{lightyellow}{0.630} & \colorbox{lightred}{27.47} & \colorbox{lightred}{0.856} & \colorbox{lightred}{0.244} & \colorbox{lightyellow}{70m38s}
        & 2.700 & \colorbox{lightorange}{23.77} & \colorbox{lightred}{0.771} & \colorbox{lightred}{0.340} & 248m48s
        & \colorbox{lightyellow}{17.700} & \colorbox{lightorange}{20.87} & \colorbox{lightorange}{0.741} & \colorbox{lightred}{0.221} & 646m15s
        & \colorbox{lightyellow}{0.410} & 26.00 & 0.887 & 0.178 & \colorbox{lightorange}{30m39s} \\
        \hline

        OceanXL (ours)
        & \colorbox{lightred}{0.317} & \colorbox{lightred}{25.03} & \colorbox{lightorange}{0.746} & \colorbox{lightyellow}{0.428} & \colorbox{lightred}{32m50s}
        & \colorbox{lightred}{0.204} & \colorbox{lightorange}{27.35} & \colorbox{lightorange}{0.848} & \colorbox{lightorange}{0.273} & \colorbox{lightred}{25m1s}
        & \colorbox{lightyellow}{0.545} & \colorbox{lightred}{24.29} & \colorbox{lightorange}{0.760} & \colorbox{lightorange}{0.359} & \colorbox{lightorange}{62m47s}
        & \colorbox{lightorange}{8.440} & \colorbox{lightred}{20.91} & \colorbox{lightred}{0.744} & \colorbox{lightorange}{0.227} & \colorbox{lightred}{307m20s}
        & \colorbox{lightred}{0.166} & \colorbox{lightred}{28.45} & \colorbox{lightred}{0.922} & \colorbox{lightred}{0.144} & \colorbox{lightred}{19m7s} \\

        \hline
    \end{tabular}
    \label{tab:main-result}
\end{table*}
}{\begin{table*}[t]
    \centering
    \setlength\tabcolsep{2pt}
    \captionsetup{singlelinecheck=false}
    \caption{Quantitative evaluation of the proposed approach compared with previous baselines on five large-scale underwater scenes: ISRO, KWAJ, Eiffel Tower 2016, Tabuhan P1, and Komodo. $\uparrow$ indicates higher is better, while $\downarrow$ indicates lower is better. Model size is reported in GB. The \colorbox{lightred}{best}, \colorbox{lightorange}{second best}, and \colorbox{lightyellow}{third best} results are highlighted. UW-GS$^{\ast}$ denotes a modified version combined with BSP. OOM indicates out-of-memory failure.} \vspace{-2mm}
    \resizebox{\textwidth}{!}{
    \begin{tabular}{r | ccccr | ccccr | ccccr | ccccr | ccccr}
        \toprule
        \multirow{2}{*}{Method}
        & \multicolumn{5}{c}{ISRO}
        & \multicolumn{5}{c}{KWAJ}
        & \multicolumn{5}{c}{Eiffel Tower 2016}
        & \multicolumn{5}{c}{Tabuhan P1}
        & \multicolumn{5}{c}{Komodo} \\
        \cmidrule(l{2pt}r{2pt}){2-6}
        \cmidrule(l{2pt}r{2pt}){7-11}
        \cmidrule(l{2pt}r{2pt}){12-16}
        \cmidrule(l{2pt}r{2pt}){17-21}
        \cmidrule(l{2pt}r{2pt}){22-26}

        & Size$\downarrow$ & PSNR$\uparrow$ & SSIM$\uparrow$ & LPIPS$\downarrow$ & Time$\downarrow$
        & Size$\downarrow$ & PSNR$\uparrow$ & SSIM$\uparrow$ & LPIPS$\downarrow$ & Time$\downarrow$
        & Size$\downarrow$ & PSNR$\uparrow$ & SSIM$\uparrow$ & LPIPS$\downarrow$ & Time$\downarrow$
        & Size$\downarrow$ & PSNR$\uparrow$ & SSIM$\uparrow$ & LPIPS$\downarrow$ & Time$\downarrow$
        & Size$\downarrow$ & PSNR$\uparrow$ & SSIM$\uparrow$ & LPIPS$\downarrow$ & Time$\downarrow$ \\
        \midrule

        UW-GS$^{\ast}$ (+BSP)
        & \colorbox{lightorange}{0.923} & 20.53 & 0.669 & 0.442 & 245m45s
        & 1.070 & 21.32 & 0.677 & 0.448 & 198m22s
        & 1.320 & 19.98 & 0.592 & 0.514 & 342m17s
        & OOM & OOM & OOM & OOM & OOM
        & 0.891 & 22.56 & 0.634 & 0.199 & 87m43s \\

        Hierarchical-GS
        & 13.310 & 20.28 & 0.648 & 0.452 & 320m8s
        & 6.280 & 21.42 & 0.679 & \colorbox{lightyellow}{0.380} & 115m18s
        & 9.110 & 17.30 & 0.519 & 0.462 & \colorbox{lightyellow}{180m56s}
        & 39.530 & 16.28 & \colorbox{lightyellow}{0.404} & \colorbox{lightyellow}{0.395} & \colorbox{lightyellow}{561m50s}
        & 1.590 & 23.98 & 0.796 & 0.184 & 55m51s \\

        CityGaussian
        & 2.050 & 21.42 & 0.695 & 0.490 & \colorbox{lightyellow}{150m10s}
        & 1.090 & 21.90 & 0.699 & 0.405 & 109m27s
        & \colorbox{lightorange}{0.450} & 21.91 & \colorbox{lightyellow}{0.700} & \colorbox{lightyellow}{0.425} & 426m0s
        & \colorbox{lightred}{3.800} & \colorbox{lightyellow}{18.31} & \colorbox{lightyellow}{0.568} & 0.417 & \colorbox{lightorange}{430m12s}
        & 2.440 & \colorbox{lightyellow}{28.09} & \colorbox{lightyellow}{0.905} & \colorbox{lightyellow}{0.163} & 432m49s \\

        Octree-GS
        & \colorbox{lightyellow}{1.420} & \colorbox{lightyellow}{24.14} & \colorbox{lightred}{0.753} & \colorbox{lightred}{0.377} & \colorbox{lightorange}{52m17s}
        & \colorbox{lightorange}{0.373} & \colorbox{lightyellow}{24.20} & \colorbox{lightyellow}{0.724} & \colorbox{lightyellow}{0.358} & \colorbox{lightorange}{37m41s}
        & \colorbox{lightred}{0.184} & \colorbox{lightyellow}{23.20} & 0.671 & 0.444 & \colorbox{lightred}{26m22s}
        & OOM & OOM & OOM & OOM & OOM
        & \colorbox{lightorange}{0.181} & \colorbox{lightorange}{28.31} & \colorbox{lightorange}{0.921} & \colorbox{lightorange}{0.149} & \colorbox{lightyellow}{52m22s} \\

        BlockGaussian
        & 3.300 & \colorbox{lightorange}{24.43} & \colorbox{lightyellow}{0.729} & \colorbox{lightorange}{0.404} & 269m4s
        & \colorbox{lightyellow}{0.630} & \colorbox{lightred}{27.47} & \colorbox{lightred}{0.856} & \colorbox{lightred}{0.244} & \colorbox{lightyellow}{70m38s}
        & 2.700 & \colorbox{lightorange}{23.77} & \colorbox{lightred}{0.771} & \colorbox{lightred}{0.340} & 248m48s
        & \colorbox{lightyellow}{17.700} & \colorbox{lightorange}{20.87} & \colorbox{lightorange}{0.741} & \colorbox{lightred}{0.221} & 646m15s
        & \colorbox{lightyellow}{0.410} & 26.00 & 0.887 & 0.178 & \colorbox{lightorange}{30m39s} \\
        \hline

        OceanXL (ours)
        & \colorbox{lightred}{0.317} & \colorbox{lightred}{25.03} & \colorbox{lightorange}{0.746} & \colorbox{lightyellow}{0.428} & \colorbox{lightred}{32m50s}
        & \colorbox{lightred}{0.204} & \colorbox{lightorange}{27.35} & \colorbox{lightorange}{0.848} & \colorbox{lightorange}{0.273} & \colorbox{lightred}{25m1s}
        & \colorbox{lightyellow}{0.545} & \colorbox{lightred}{24.29} & \colorbox{lightorange}{0.760} & \colorbox{lightorange}{0.359} & \colorbox{lightorange}{62m47s}
        & \colorbox{lightorange}{8.440} & \colorbox{lightred}{20.91} & \colorbox{lightred}{0.744} & \colorbox{lightorange}{0.227} & \colorbox{lightred}{307m20s}
        & \colorbox{lightred}{0.166} & \colorbox{lightred}{28.45} & \colorbox{lightred}{0.922} & \colorbox{lightred}{0.144} & \colorbox{lightred}{19m7s} \\

        \bottomrule
    \end{tabular}
    }
    \label{tab:main-result}
\end{table*}
}

\subsection{Results and Comparison}
\subsubsection{Partitioning Results}
The partitioning results obtained on the ISRO scene using the proposed BSP strategy are shown in~\autoref{fig:Partitioning}. In contrast to the BlockGaussian partitioning strategy~\cite{wu2025blockgaussian}, shown on the left, which mechanically splits the scene into two parts at the midpoint, our method introduces unaligned partition boundaries to satisfy a maximum point budget determined by the available GPU capacity. In this experiment, the point budget is set to 40,000 points per partition. As a result, our strategy achieves a more balanced point-cloud distribution across partitions. 

\subsubsection{Quantitative Comparison} 
~\autoref{tab:main-result} reports the quantitative comparison between the proposed method and existing large-scale Gaussian splatting approaches. OceanXL achieves the best efficiency, producing the smallest model sizes and fastest training times across most datasets, while using only 32.3\% as many primitives as BlockGaussian on average. In terms of PSNR, SSIM, and LPIPS, the proposed method performs comparably to or better than existing state-of-the-art methods, demonstrating a strong efficiency--quality trade-off for large-scale underwater reconstruction. There are two notable exceptions. On the Eiffel Tower 2016 scene, Octree-GS achieves a smaller model size and faster training speed, while OceanXL produces higher reconstruction quality. On the Tabuhan P1 scene, OceanXL yields a larger model than CityGaussian; however, we restrict CityGaussian point-cloud growth to ensure all methods remain trainable under the single-GPU setting.

\subsubsection{Qualitative Comparison}
\begin{figure}
    \centering
    \includegraphics[width=\linewidth]{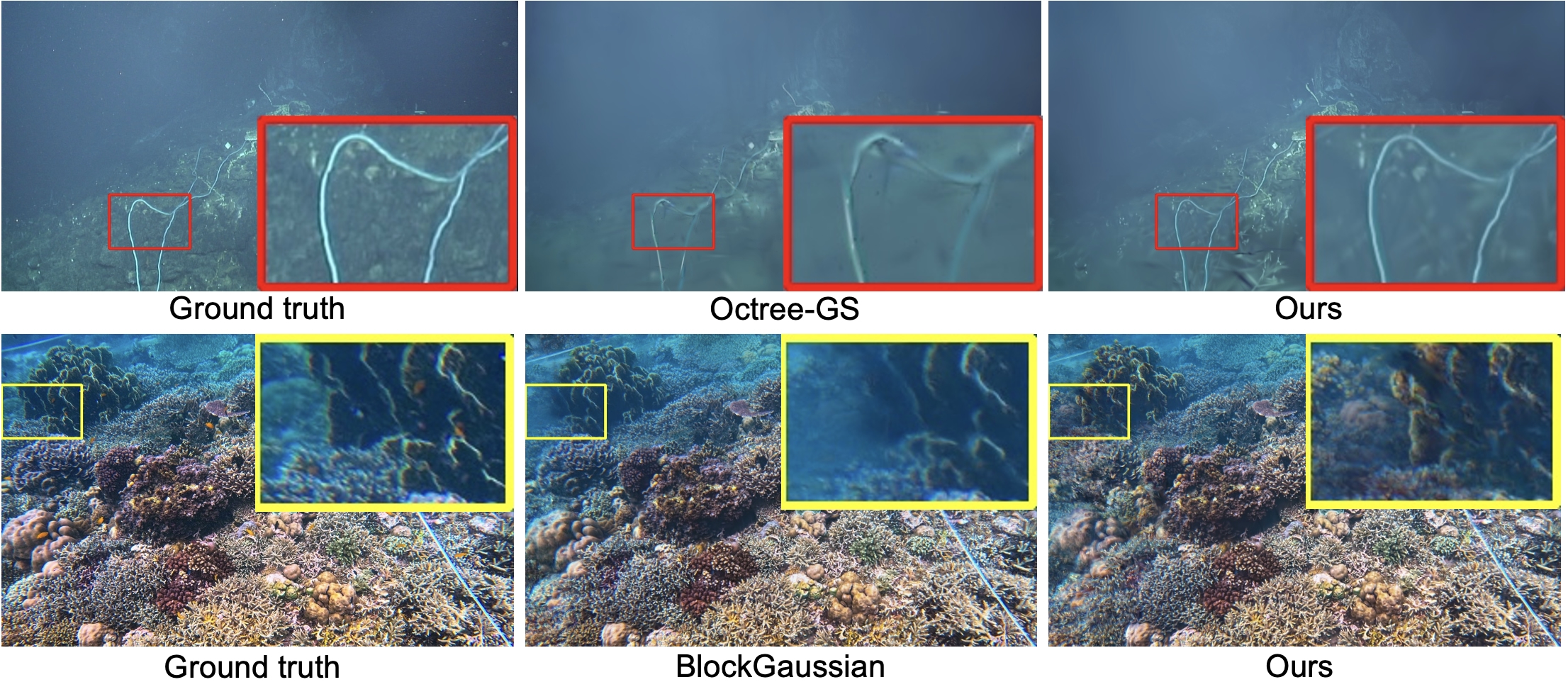} 
    \caption{Qualitative comparison of the proposed method and baseline approaches on (top) Eiffel Tower 2016 and (bottom) Tabuhan P1 scenes.}
    \Description{Qualitative rendering comparison between OceanXL and baseline methods on the Eiffel Tower 2016 and Tabuhan P1 scenes.}
    \label{fig:results}
\end{figure}

~\autoref{fig:results} presents a qualitative comparison between the proposed method and baseline approaches. OceanXL recovers fine underwater structures while maintaining a compact scene representation. Additional qualitative comparisons are provided in ~\autoref{fig:oceanxresults} and ~\autoref{fig:comparion}, while point cloud visualizations are shown in ~\autoref{fig:points_others}.

\subsection{Ablation Study}
To validate the contribution of each component, we conduct the ablation studies reported in~\autoref{tab:ablation}. We use 3DGS with the BlockGaussian partitioning strategy as the baseline. Replacing it with the proposed Balanced Scene Partitioning (BSP) (V1) improves both reconstruction quality and training efficiency. This gain is attributed to the more balanced distribution of geometry across partitions, which reduces the number of reconstruction chunks
and improves single-GPU optimization efficiency. Incorporating Underwater-Compensated Densification (UCD) (V2) further improves PSNR, SSIM, and LPIPS by compensating for attenuation-induced densification failures in degraded underwater regions, although it increases the model size. Introducing Wise Adaptive Pruning (WAP) without DoG activation (V3) substantially reduces model size and training time, but degrades reconstruction quality. Activating DoG components for all remaining primitives (V4) recovers quality, but introduces computational overhead. In contrast, OceanXL combines WAP with selective DoG activation for reconstruction-sensitive primitives. Although selective activation slightly decreases reconstruction metrics compared with V2 and V4, it achieves a more favorable trade-off between reconstruction quality, compactness, and efficiency. 
\aptLtoX{\begin{table}
    \centering
    \footnotesize
    \setlength{\tabcolsep}{3pt}
    \captionsetup{singlelinecheck=false}
    \vspace{-1mm}
    \caption{The reconstruction performance of different variants.
    We report the average PSNR, SSIM, and LPIPS scores for the
    \textbf{Abyssal dataset} with 3DGS as the backbone.
    Full denotes full DoG activation, and selective denotes the proposed
    selective DoG activation strategy. 3DGS$^{\ast}$ denotes a modified version combined with BlockGaussian partitioning strategy} \vspace{-1mm}

    \resizebox{\columnwidth}{!}{
    \begin{tabular}{r|ccccc|ccc|cc}
    \toprule
    \multicolumn{1}{l}{}
    & \multicolumn{5}{c|}{\textbf{Components}}
    & \multicolumn{3}{c|}{\textbf{Reconstruction Metrics}}
    & \multicolumn{2}{c}{\textbf{Efficiency Metrics}} \\

    Variant
    & BSP
    & UCD
    & WAP
    & Full 
    & Selective
    & PSNR$\uparrow$
    & SSIM$\uparrow$
    & LPIPS$\downarrow$
    & Time$\downarrow$
    & Size$\downarrow$ \\
    \midrule

3DGS$^{\ast}$
& \xmark & \xmark & \xmark & \xmark & \xmark
& 22.40
& 0.719
& 0.420
& 43m7s
& 0.670 \\

V1
& \cmark & \xmark & \xmark & \xmark & \xmark
& 23.05
& 0.721
& 0.402
& \colorbox{lightyellow}{31m17s}
& 0.593 \\

V2
& \cmark & \cmark & \xmark & \xmark & \xmark
& \colorbox{lightred}{26.36}
& \colorbox{lightred}{0.818}
& \colorbox{lightred}{0.334}
& 37m13s
& 0.768 \\

V3
& \cmark & \cmark & \cmark & \xmark & \xmark
& 26.04
& 0.788
& 0.361
& \colorbox{lightred}{25m30s}
& \colorbox{lightred}{0.242} \\

V4
& \cmark & \cmark & \cmark & \cmark & \xmark
& \colorbox{lightorange}{26.26}
& \colorbox{lightorange}{0.809}
& \colorbox{lightorange}{0.344}
& 37m12s
& \colorbox{lightyellow}{0.264} \\
\midrule

OceanXL (ours)
& \cmark & \cmark & \cmark & \xmark & \cmark
& \colorbox{lightyellow}{26.19}
& \colorbox{lightyellow}{0.797}
& \colorbox{lightyellow}{0.351}
& \colorbox{lightorange}{28m56s}
& \colorbox{lightorange}{0.261} \\
    \bottomrule
    \end{tabular}
    }

   % \vspace{-1mm}
    \label{tab:ablation}
%\end{table}
\end{table}

\begin{table}
    \centering
    \footnotesize
    \setlength{\tabcolsep}{2pt}
    \captionsetup{singlelinecheck=false}

    {
    \caption{
    Quantitative comparisons on the small-scale \textbf{SeaThru-NeRF} dataset.
    $\uparrow$ indicates higher is better, while $\downarrow$ indicates lower is better.
    Model size is reported in GB.
    } \vspace{-1mm}

    %\begin{adjustbox}{max width=\columnwidth}
    \begin{tabular}{r|c|ccc}
        \toprule
        Method
        & Size$\downarrow$
        & PSNR$\uparrow$
        & SSIM$\uparrow$
        & LPIPS$\downarrow$ \\
        \midrule

        WaterSplatting
        & \colorbox{lightorange}{0.689}
        & 28.75 & 0.904 & 0.148 \\

        SeaSplat
        & 4.000
        & 27.11 & 0.885 & 0.183 \\

        GaussianSplashing
        & \colorbox{lightyellow}{0.963}
        & \colorbox{lightorange}{29.00}
        & 0.910
        & \colorbox{lightorange}{0.145} \\

        RUSplatting
        & 1.530
        & \colorbox{lightred}{29.29}
        & \colorbox{lightorange}{0.916}
        & 0.166 \\

        UW-GS
        & 1.160
        & \colorbox{lightyellow}{28.82}
        & \colorbox{lightred}{0.918}
        & \colorbox{lightred}{0.143} \\
        \midrule

        OceanXL (ours)
        & \colorbox{lightred}{0.125}
        & 28.75
        & \colorbox{lightyellow}{0.912}
        & \colorbox{lightyellow}{0.146} \\
        \bottomrule
    \end{tabular}
    %\end{adjustbox}
    }
\vspace{-4mm}
    \label{tab:seathru-nerf-result}
\end{table}
}{\begin{table}
    \centering
    \footnotesize
    \setlength{\tabcolsep}{3pt}
    \captionsetup{singlelinecheck=false}
    \vspace{-1mm}
    \caption{The reconstruction performance of different variants.
    We report the average PSNR, SSIM, and LPIPS scores for the
    \textbf{Abyssal dataset} with 3DGS as the backbone.
    Full denotes full DoG activation, and selective denotes the proposed
    selective DoG activation strategy. 3DGS$^{\ast}$ denotes a modified version combined with BlockGaussian partitioning strategy} \vspace{-1mm}

    \resizebox{\columnwidth}{!}{
    \begin{tabular}{r|ccccc|ccc|cc}
    \toprule
    \multicolumn{1}{l}{}
    & \multicolumn{5}{c|}{\textbf{Components}}
    & \multicolumn{3}{c|}{\textbf{Reconstruction Metrics}}
    & \multicolumn{2}{c}{\textbf{Efficiency Metrics}} \\

    Variant
    & BSP
    & UCD
    & WAP
    & Full 
    & Selective
    & PSNR$\uparrow$
    & SSIM$\uparrow$
    & LPIPS$\downarrow$
    & Time$\downarrow$
    & Size$\downarrow$ \\
    \midrule

3DGS$^{\ast}$
& \xmark & \xmark & \xmark & \xmark & \xmark
& 22.40
& 0.719
& 0.420
& 43m7s
& 0.670 \\

V1
& \cmark & \xmark & \xmark & \xmark & \xmark
& 23.05
& 0.721
& 0.402
& \colorbox{lightyellow}{31m17s}
& 0.593 \\

V2
& \cmark & \cmark & \xmark & \xmark & \xmark
& \colorbox{lightred}{26.36}
& \colorbox{lightred}{0.818}
& \colorbox{lightred}{0.334}
& 37m13s
& 0.768 \\

V3
& \cmark & \cmark & \cmark & \xmark & \xmark
& 26.04
& 0.788
& 0.361
& \colorbox{lightred}{25m30s}
& \colorbox{lightred}{0.242} \\

V4
& \cmark & \cmark & \cmark & \cmark & \xmark
& \colorbox{lightorange}{26.26}
& \colorbox{lightorange}{0.809}
& \colorbox{lightorange}{0.344}
& 37m12s
& \colorbox{lightyellow}{0.264} \\
\midrule

OceanXL (ours)
& \cmark & \cmark & \cmark & \xmark & \cmark
& \colorbox{lightyellow}{26.19}
& \colorbox{lightyellow}{0.797}
& \colorbox{lightyellow}{0.351}
& \colorbox{lightorange}{28m56s}
& \colorbox{lightorange}{0.261} \\
    \bottomrule
    \end{tabular}
    }

    \label{tab:ablation}
%\end{table}
\vspace{3mm}
%\begin{table}[t]
    \centering
    \footnotesize
    \setlength{\tabcolsep}{2pt}
    \captionsetup{singlelinecheck=false}

    {
    \caption{
    Quantitative comparisons on the small-scale \textbf{SeaThru-NeRF} dataset.
    $\uparrow$ indicates higher is better, while $\downarrow$ indicates lower is better.
    Model size is reported in GB.
    } \vspace{-1mm}

    %\begin{adjustbox}{max width=\columnwidth}
    \begin{tabular}{r|c|ccc}
        \toprule
        Method
        & Size$\downarrow$
        & PSNR$\uparrow$
        & SSIM$\uparrow$
        & LPIPS$\downarrow$ \\
        \midrule

        WaterSplatting
        & \colorbox{lightorange}{0.689}
        & 28.75 & 0.904 & 0.148 \\

        SeaSplat
        & 4.000
        & 27.11 & 0.885 & 0.183 \\

        GaussianSplashing
        & \colorbox{lightyellow}{0.963}
        & \colorbox{lightorange}{29.00}
        & 0.910
        & \colorbox{lightorange}{0.145} \\

        RUSplatting
        & 1.530
        & \colorbox{lightred}{29.29}
        & \colorbox{lightorange}{0.916}
        & 0.166 \\

        UW-GS
        & 1.160
        & \colorbox{lightyellow}{28.82}
        & \colorbox{lightred}{0.918}
        & \colorbox{lightred}{0.143} \\
        \midrule

        OceanXL (ours)
        & \colorbox{lightred}{0.125}
        & 28.75
        & \colorbox{lightyellow}{0.912}
        & \colorbox{lightyellow}{0.146} \\
        \bottomrule
    \end{tabular}
    %\end{adjustbox}
    }
\vspace{-4mm}
    \label{tab:seathru-nerf-result}
\end{table}
}

\subsection{Generalization to Small-Scale Underwater Scenes}
Although OceanXL is primarily designed for large-scale underwater reconstruction, we further evaluate its generalization ability on the small-scale SeaThru-NeRF dataset. Unlike the large-scale baselines evaluated in~\autoref{tab:main-result}, this experiment compares OceanXL with representative underwater-specific Gaussian Splatting methods, including WaterSplatting~\cite{li2024watersplatting}, SeaSplat~\cite{yang2025seasplat}, GaussianSplashing~\cite{mualem2024gaussian}, RUSplatting~\cite{jiang2025rusplatting}, and UW-GS~\cite{wang2025uw}. We follow the standard evaluation protocol and report PSNR, SSIM, LPIPS, and model size. As shown in~\autoref{tab:seathru-nerf-result}, OceanXL achieves competitive reconstruction quality on SeaThru-NeRF while requiring only 0.125 GB of storage, corresponding to an 82–97\% reduction compared with the evaluated underwater-specific baselines. These results support the compactness and small-scale generalization of OceanXL, rather than serving as evidence of large-scale scalability.
%}
% ----------------------------------------------------
\section{Discussion}
Our results suggest that scalable underwater scene representation depends less on the raw expressivity of Gaussian primitives than on how computation, capacity, and visibility are distributed across the scene. We further discuss the resulting trade-offs, practical integration into interactive graphics pipelines, and the framework’s generalization and limitations.

\subsection{Effectiveness for Large-Scale Data}
Naive partitioning of large-scale scenes often yields imbalanced chunks: dense partitions incur high optimization costs, whereas sparse ones lack sufficient observations for stable reconstruction. BSP mitigates this imbalance by distributing partition complexity more evenly, improving optimization stability in large-scale underwater scenes, consistent with scalable rendering systems~\cite{kerbl2024hierarchical,Lee:GSScale:2026}.

Adaptive pruning further improves scalability by suppressing redundant primitives introduced during densification. Standard 3DGS tends to over-allocate Gaussians in visually ambiguous regions~\cite{bagdasarian20253dgs}, which is exacerbated underwater by scattering and attenuation. By progressively pruning low-contribution primitives and selectively activating DoG components in reconstruction-sensitive regions, our strategy allocates representational capacity more efficiently while preserving reconstruction fidelity.

\subsection{Underwater Reconstruction Quality}
Compared with previous underwater 3DGS approaches discussed in ~\autoref{sec:relatedwork}, the proposed Underwater-Compensated Densification offers a substantially lighter-weight alternative using only three depth-dependent densification weights. Despite its simplicity, ~\autoref{fig:Komodo_points} shows that OceanXL achieves improved underwater color consistency and correction compared with standard large-scene GS approaches.
However, the current formulation uses fixed depth boundaries, where $z_\text{near}$ and $z_\text{far}$ are set to $\frac{1}{3}z_\text{max}$ and $\frac{2}{3}z_\text{max}$, respectively. While effective in practice, these predefined thresholds may not fully capture spatially varying underwater optical properties. More adaptive depth stratification could further improve reconstruction quality.

\subsection{Selective Activation of DoG Primitives}
By allocating additional capacity only to reconstruction-sensitive regions, the method avoids unnecessary computation and memory overhead in low-detail areas, which is beneficial for underwater scenes with severe visibility degradation.
This design can integrate naturally with modern 3DGS pipelines, as DoG activation operates directly at the primitive level without auxiliary neural decoding, remaining compatible with real-time rasterization and chunk-based VR/AR streaming systems.
Broadly, the strategy generalizes beyond underwater environments, since it is driven by reconstruction sensitivity rather than domain-specific priors. However, the current formulation still relies on manually defined activation thresholds, which may not always align with perceptual importance under different types of degradation.

\subsection{Failure Cases}
OceanXL still struggles with sparse views, as shown in ~\autoref{fig:failcase}. When neighboring viewpoints share limited common observations, the aggressive compactness introduced by adaptive pruning may reduce the representational capacity needed to recover fine structures, leading to slightly blurrier renderings. In future work, we plan to explore sparse-view-aware pruning strategies and incorporate geometry-consistent generative priors to improve fine-detail recovery under limited view overlap while mitigating hallucinated structures. In addition, dynamic marine life is not a focus of OceanXL. Transient fish observations are often filtered out during SfM initialization or suppressed during multi-view 3DGS optimization because they are not geometrically consistent across views. Nevertheless, fish that remain visible across multiple views or move rapidly may still introduce floating, blurring, or ghosting artifacts. Since OceanXL does not employ explicit dynamic-object masks, handling such content remains a limitation. Integrating motion segmentation or dynamic-object masking is left for future work.

\subsection{Practical Advantages for Real-Time Graphics and VR Applications}
Beyond quantitative comparisons, several properties of OceanXL make it well suited for downstream graphics applications. Its block-wise representation naturally supports streaming and level-of-detail (LoD) strategies in VR and real-time engines, allowing chunks to be independently loaded, swapped, or culled. The compact per-chunk footprint, typically a few hundred megabytes even for large marine scenes, remains practical for consumer GPUs and head-mounted displays \cite{zhao2026clm}, unlike many prior large-scene methods. In addition, the lightweight density control requires no auxiliary network during inference, enabling straightforward integration into existing splatting-based rasterizers and 3DGS-style pipelines commonly used in film and game production. Together, these properties support neural rendering as a reusable asset format rather than a one-off reconstruction artifact.

%\vspace{1mm}
\subsection{Generalization Beyond Underwater Scenes}
The underwater-specific component, such as depth-banded densification compensation, can also generalize to other visibility-degraded environments \cite{li2017haze}, such as foggy outdoor scenes \cite{ramazzina2023scatternerf, yu2026dehazegs}, smoke-filled architectural spaces \cite{zheng20263d}, and atmospheric VFX backgrounds where visibility progressively decays with distance.

\section{Conclusion}

We presented OceanXL, a scalable framework for large-scale underwater scene reconstruction based on 3D Gaussian Splatting. By combining BSP, underwater-aware lightweight density control, and selective Difference-of-Gaussians activation, the proposed method enables efficient and compact reconstruction of extensive underwater environments while preserving high rendering fidelity. Experimental results across multiple underwater datasets demonstrate improved efficiency-quality trade-offs compared with existing large-scale Gaussian splatting approaches.

Current limitations include fixed depth stratification and lack of explicit dynamic modeling. Future work includes uncertainty-aware merging, physics-informed partitioning, and extension to temporal reconstruction. By making large-scale underwater 3D reconstruction computationally accessible, OceanXL opens new possibilities for marine exploration, archaeological documentation, and immersive ocean experiences.

\begin{acks}
This work was supported by the EPSRC ECR International Collaboration Grants
(EP/Y002490/1) and the UKRI MyWorld Strength in Places Programme (SIPF00006/1).
We would like to thank Rachel Fu and Lionel Taillens from the OceanX media team
for providing the raw videos and metadata for the \textbf{OceanXplore} dataset.
\end{acks}
\newpage
\bibliographystyle{ACM-Reference-Format}
\bibliography{main}

\appendix
% ---------- figure-only pages start here ----------------
\clearpage
\begin{figure}
  \centering
   \includegraphics[width=1\linewidth]{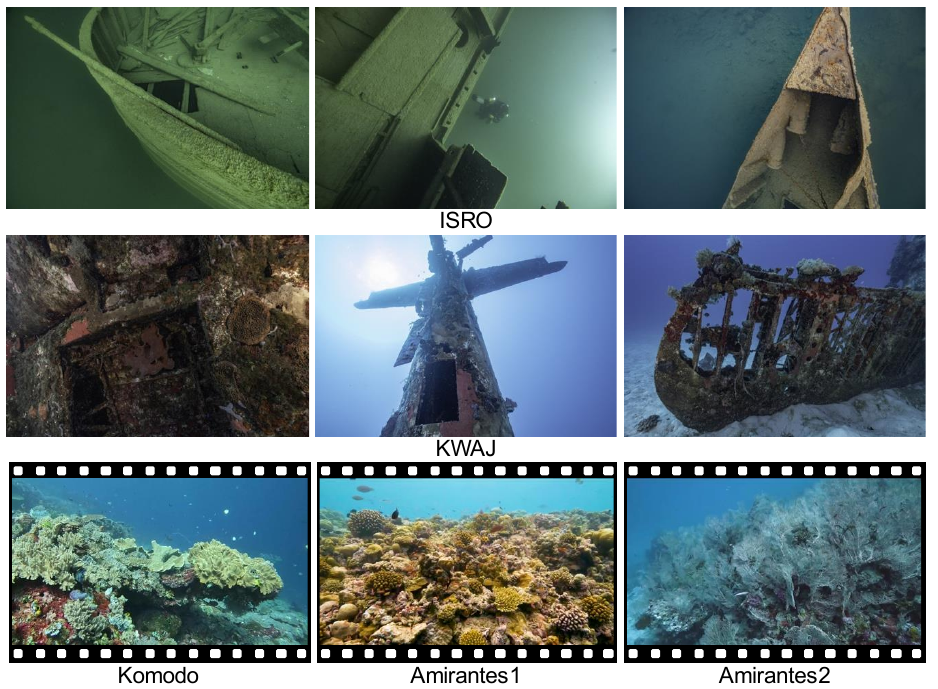}
   \caption{Samples from the Abyssal and OceanXplore datasets.. (Top) Sample images from \textbf{ISRO} Scene. (Middle) Sample images from \textbf{KWAJ} Scene. (Bottom) Sample frames from \textbf{Komodo}, \textbf{Amirantes1}, and \textbf{Amirantes2} Scene.}
   \Description{XXX} \vspace{-3mm}
   \label{fig:Abyssal-Artifacts}
\end{figure}

\begin{figure}
  \centering
   \includegraphics[width=1.0\linewidth]{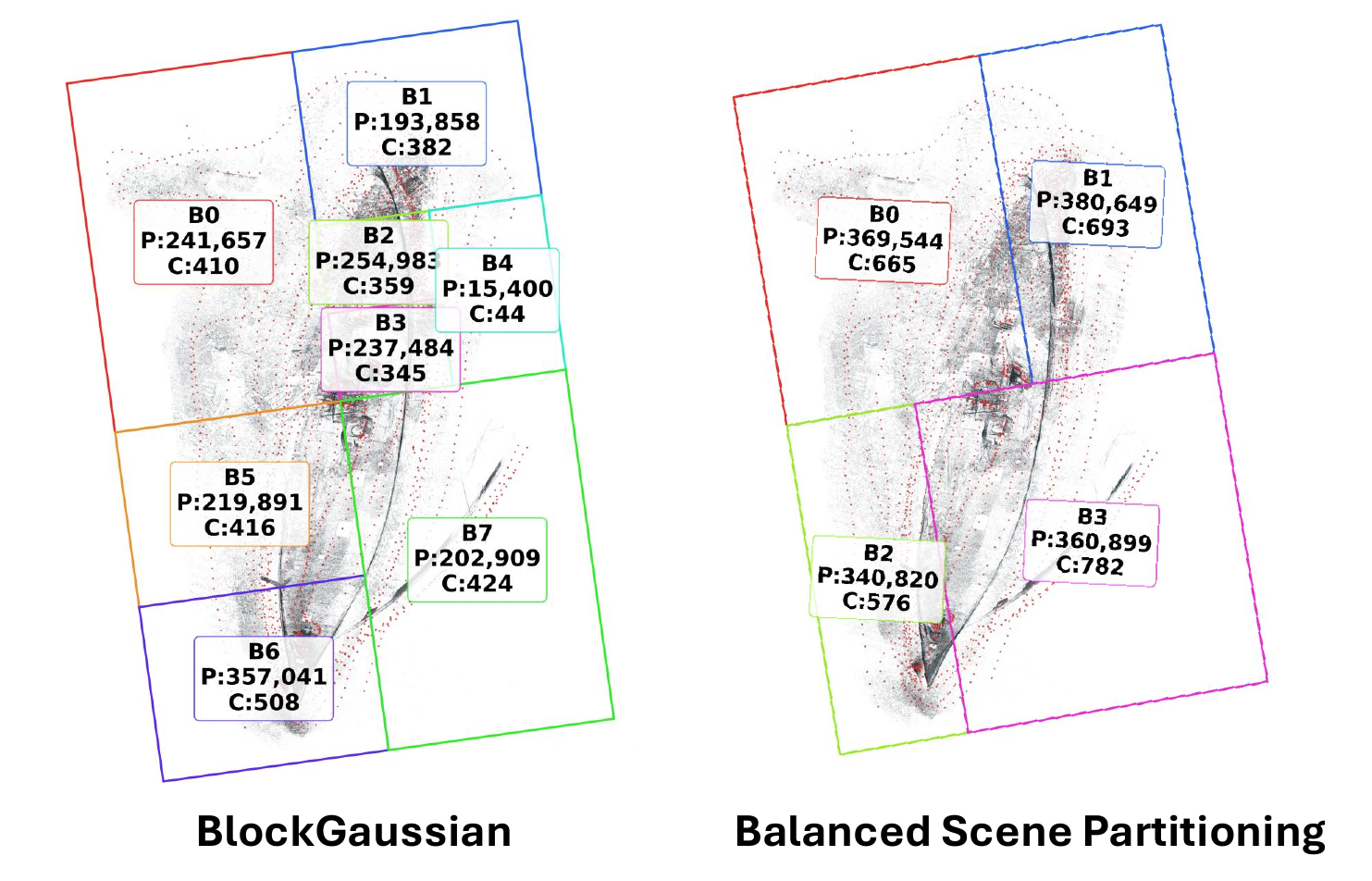}
   \caption{\textbf{Partitioning comparison on a large-scale underwater scene.} BlockGaussian (left) produces imbalanced partitions with large variations in SfM point counts and camera distributions across chunks. In contrast, the proposed BSP strategy (right) generates more balanced partitions, leading to improved workload distribution for large-scale reconstruction. P denotes the number of SfM points and C denotes the number of cameras. }
   \Description{XXX}
   \label{fig:Partitioning}
\end{figure}

\begin{figure}
    \centering
    \includegraphics[width=\columnwidth]{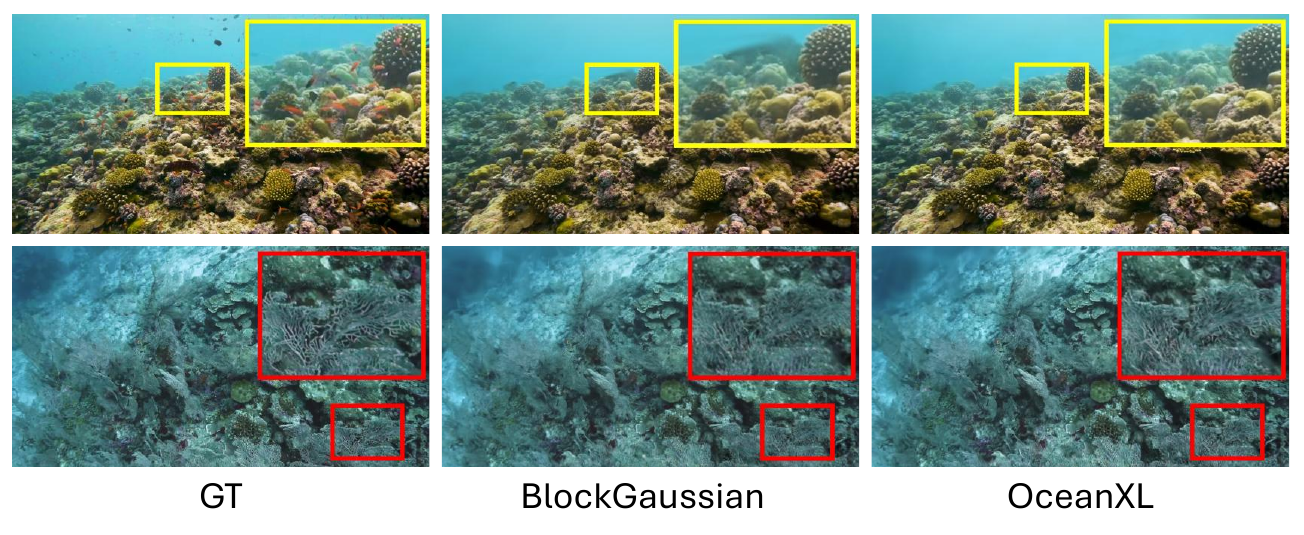}
    \caption{\textbf{Novel view rendering comparison with the baselines. Top: \textbf{Amirantes1}. Bottom: \textbf{Amirantes2}.
    Best viewed zoomed in.}}
    \Description{XXX}
    \label{fig:oceanxresults}
\end{figure}

\begin{figure}
    \centering
    \includegraphics[width=\columnwidth]{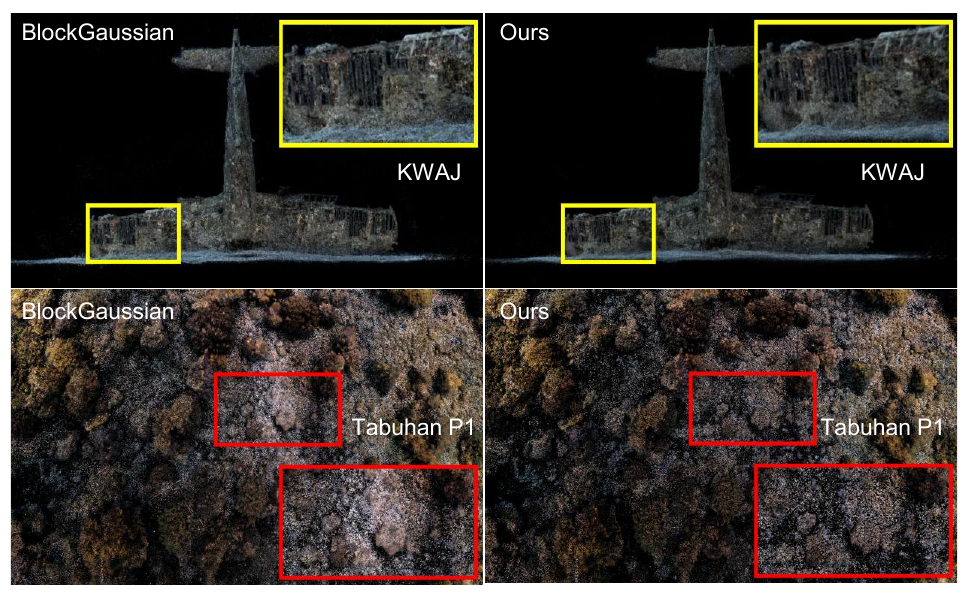}
    \caption{\textbf{Point cloud comparison of KWAJ and Tabuhan P1 scenes between BlockGaussian (left) and our OceanXL (right).
    Best viewed zoomed in.}}
    \Description{XXX}
    \label{fig:points_others}
\end{figure}

\begin{figure}
    \centering
    \includegraphics[width=\columnwidth]{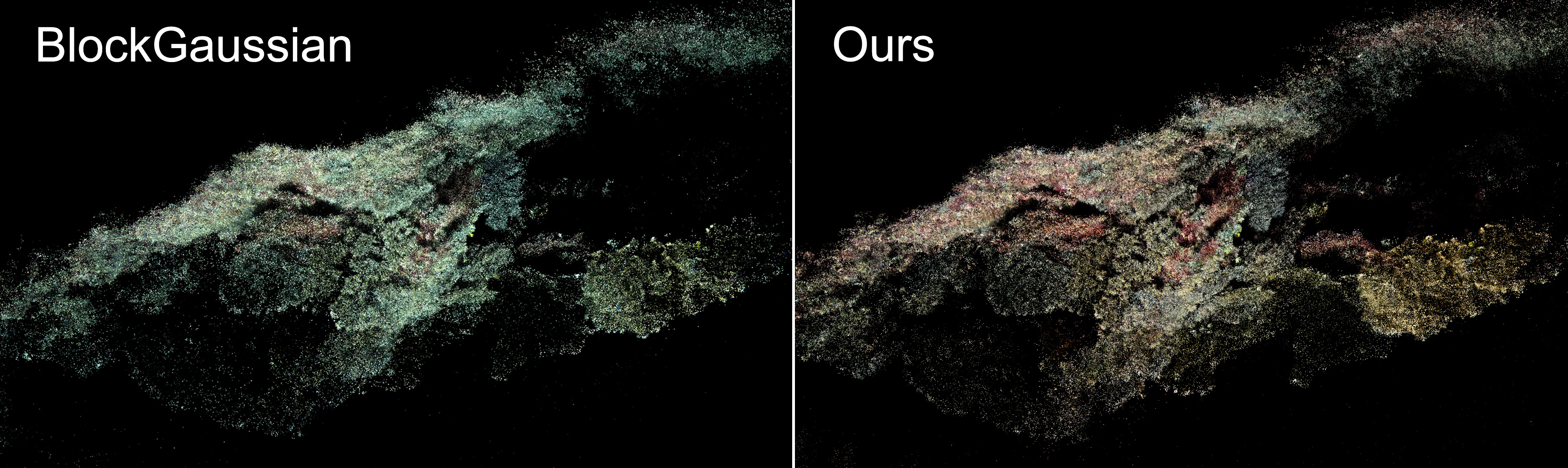}
    \caption{Underwater color correction on the \textbf{Komodo} scene.
    Compared with the raw point cloud colors produced by BlockGaussian
    (left), OceanXL (right) achieves more natural color restoration.}
    \Description{XXX}
    \label{fig:Komodo_points}
\end{figure}

\begin{figure}
    \centering
    \includegraphics[width=\columnwidth]{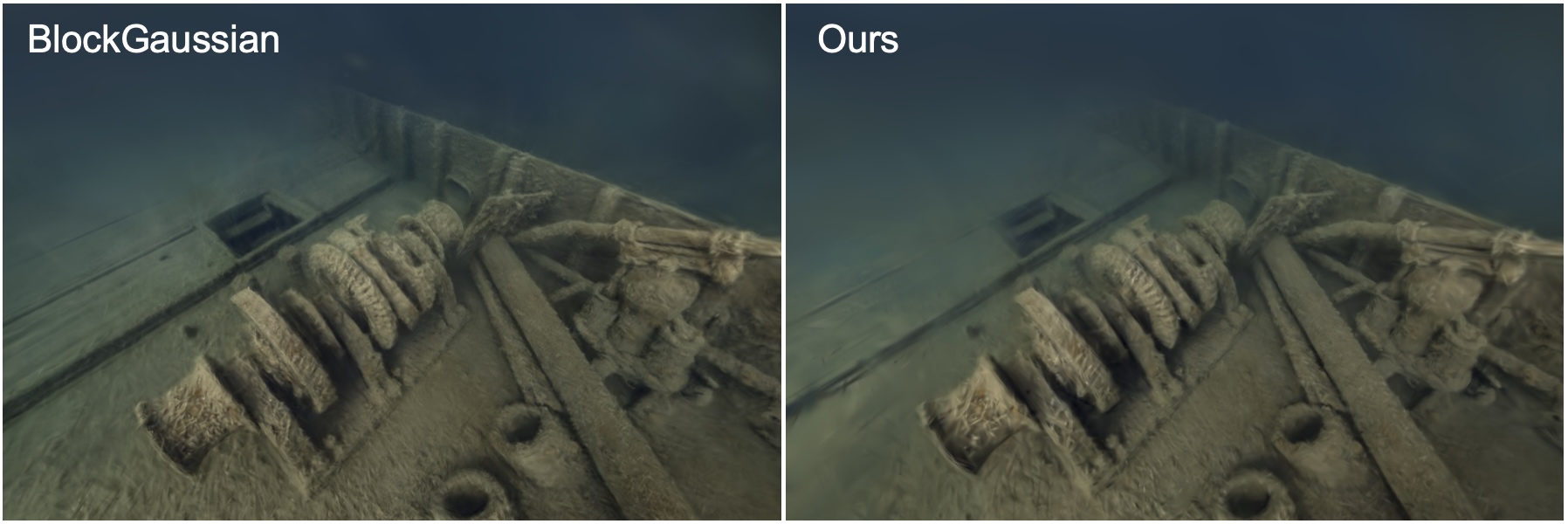}
    \caption{\textbf{Failure case.} When overlap between nearby views is
    limited, OceanXL may produce blurrier renderings
    (e.g., in some part of \textbf{KWAJ}), as shown on the right,
    compared with BlockGaussian (left), which preserves finer details.
    However, BlockGaussian requires substantially more Gaussian primitives
    to achieve this quality.}
    \Description{XXX}
    \label{fig:failcase}
\end{figure}

\begin{figure}
    \includegraphics[width=1.0\columnwidth]{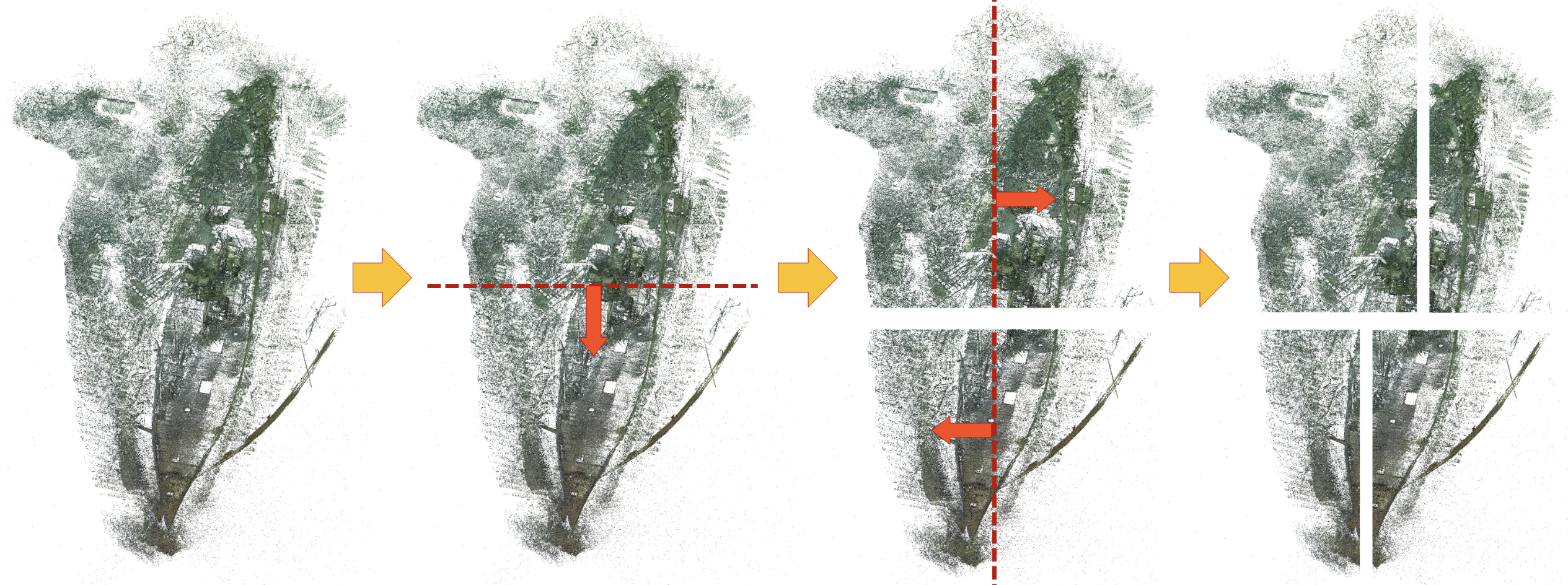}
    \caption{\textbf{Example of the proposed Balanced Scene Partitioning on the ISRO dataset.} The \textcolor{red}{red dashed line} indicates the splitting boundary initialized at the midpoint of the longest axis, while the \textcolor{red}{red arrows} denote the search directions used to balance the point counts of the two partitions.}
    \Description{XXX} \vspace{-3mm}
    \label{fig:chunking}
\end{figure}

\begin{figure*}
\vspace{-2mm}
    \centering
    \includegraphics[width=\textwidth]{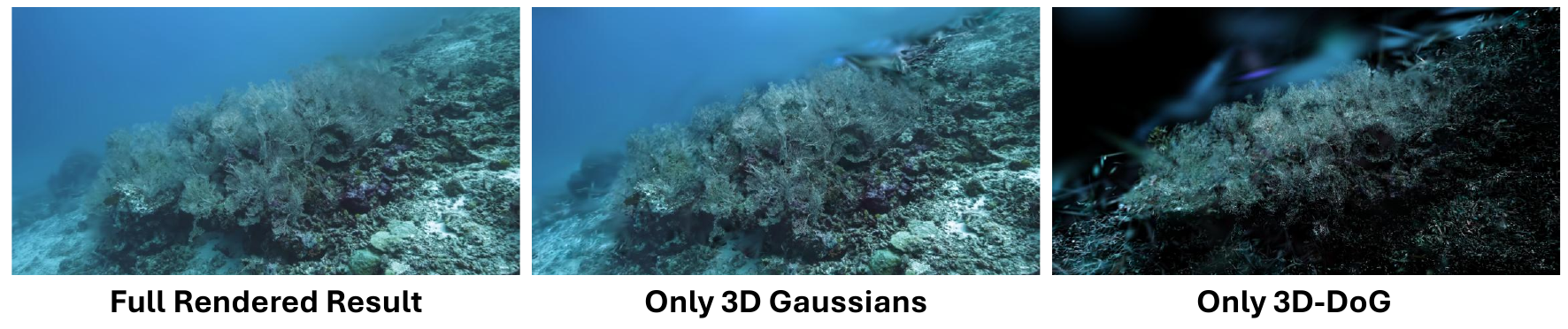}
    \caption{\textbf{3D-DoG comparison.}
    Rendering quality comparison (\textbf{Amirantes2} scene) between
    OceanXL using both 3DGS and 3D-DoG primitives (left) and using only
    standard 3DGS primitives (middle). The results show that 3D-DoG
    primitives improve high-frequency reconstruction and fine structural
    detail recovery. The right image visualizes the regions where
    3D-DoG primitives are activated.}
    \Description{XXX}
    \label{fig:DoG}

\end{figure*}

\begin{figure*}
    \centering
    \includegraphics[width=0.95\textwidth]{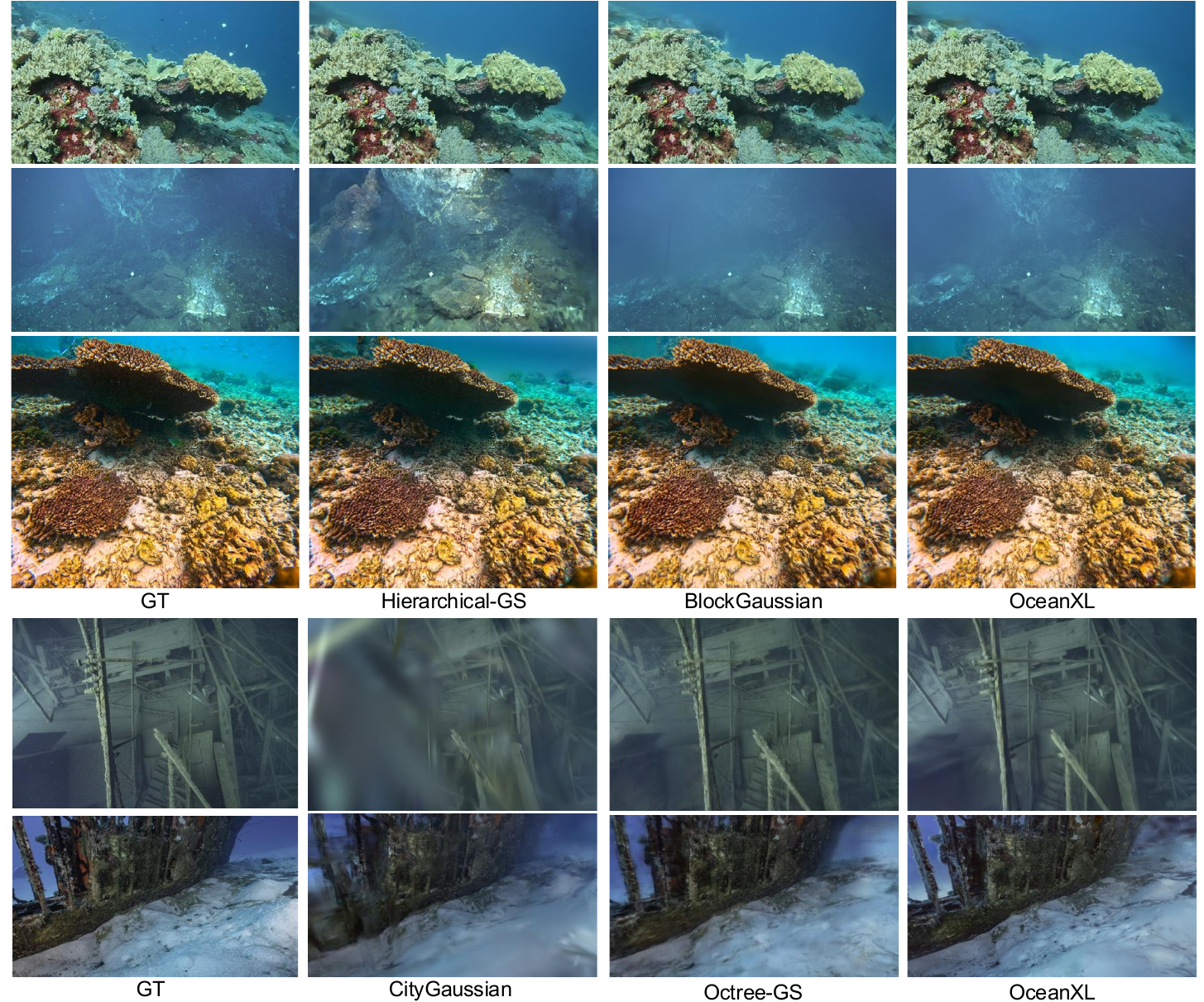}
    \caption{\textbf{Novel view rendering comparison with baseline methods.}
    From top to bottom: \textbf{Komodo}, \textbf{Eiffel Tower 2016},
    \textbf{Tabuhan P1}, \textbf{ISRO}, and \textbf{KWAJ}.
    Best viewed zoomed in. Overall, OceanXL achieves a better balance
    between reconstruction fidelity and model compactness, preserving
    fine underwater structures while maintaining efficient representations.
    We also include a representative case from \textbf{Eiffel Tower 2016}
    where Hierarchical-GS produces visually sharper results. However,
    as shown in ~\autoref{tab:main-result}, its overall quantitative
    performance is worse than several methods, including ours, while
    requiring substantially larger model sizes.}
    \Description{Novel-view rendering comparison of OceanXL and baseline methods across the Komodo, Eiffel Tower 2016, Tabuhan P1, ISRO, and KWAJ scenes.}
    \label{fig:comparion}
    \vspace{-2mm}
\end{figure*}
\clearpage

\end{document}